\documentclass[letterpaper]{article} 
\usepackage{aaai24}  
\usepackage{times}  
\usepackage{helvet}  
\usepackage{courier}  
\usepackage[hyphens]{url}  
\usepackage{graphicx} 
\usepackage{natbib}  
\usepackage{caption} 
\usepackage{algorithm}
\usepackage{algorithmic}
\usepackage{booktabs} 
\usepackage{color}
\usepackage{amssymb}
\usepackage{amsmath}
\usepackage{multirow}

\usepackage{newfloat}
\usepackage{listings}
\DeclareCaptionStyle{ruled}{labelfont=normalfont,labelsep=colon,strut=off} 
\floatstyle{ruled}
\newfloat{listing}{tb}{lst}{}
\floatname{listing}{Listing}
\nocopyright

\title{C2G2: Controllable Co-speech Gesture Generation with Latent Diffusion Model}

\author {
    Longbin Ji\textsuperscript{\rm 1,\rm 2},
    Pengfei Wei\textsuperscript{\rm 2},
    Yi Ren\textsuperscript{\rm 2},
    Jinglin Liu\textsuperscript{\rm 2},
    Chen Zhang\textsuperscript{\rm 2},
    Xiang Yin\textsuperscript{\rm 2}
}
\affiliations {
    \textsuperscript{\rm 1}Xi'an Jiaotong Liverpool University\\
    \textsuperscript{\rm 2}Bytedance\\
    longbin.ji19@student.xjtlu.edu.cn, \{wei.pengfei, ren.yi, liu.jinglin, zhangchen.990620, yinxiang.stephen\}@bytedance.com
}

\usepackage{bibentry}

\begin{document}

\maketitle

\begin{abstract}
Co-speech gesture generation is crucial for automatic digital avatar animation. 
However, existing methods suffer from issues such as unstable training and temporal inconsistency, particularly in generating high-fidelity and comprehensive gestures. 
Additionally, these methods lack effective control over speaker identity and temporal editing of the generated gestures.
Focusing on capturing temporal latent information and applying practical controlling, we propose a \textit{Controllable Co-speech Gesture Generation} framework, named \textbf{C2G2}.
Specifically, we propose a two-stage temporal dependency enhancement strategy motivated by latent diffusion models.
We further introduce two key features to C2G2, namely a speaker-specific decoder to generate speaker-related real-length skeletons and a repainting strategy for flexible gesture generation/editing. 
Extensive experiments on benchmark gesture datasets verify the effectiveness of our proposed C2G2 compared with several state-of-the-art baselines.
The link of the project demo page can be found at \url{https://c2g2-gesture.github.io/c2_gesture}.

\end{abstract}

\section{Introduction}
Gestures that accompany speech play an essential role in human communication, providing significant support for expressing personality, emotions, and motivations of speakers.
Existing studies \cite{goldin2005hearing,goldin1999role,nyatsanga2023comprehensive} have shed light on the functions of gestures including the ability to convey motions with more communicative information and prevent excessive redundancy through mimetic expression.
As a result, the generation of appropriate co-speech gesture behaviors using AI models is one of the crucial branches in achieving Artificial General Intelligence (AGI).


Existing co-speech gesture generation methods can be broadly divided into two categories, namely rule-based and data-driven approaches. 
Early works \cite{cassell1994animated,huang2012robot} predominantly fall under the rule-based category, where gestures are selected from a pre-built gesture pool based on specifically designed heuristics. 
However, these methods often lack diversity in the generated gestures due to their heavy reliance on predefined rules.
In contrast, recent research has shifted the focus towards data-driven approaches, aiming to learn the association between speech and corresponding gestures from the wild data using deep learning-based models \cite{alexanderson2020style,ao2022rhythmic}.
Among them, GAN-based models \cite{yoon2020speech,liu2022learning} have attracted increasing attention as they can produce realistic and reliable gestures by using adversarial loss.
\begin{figure}[t]
\centering
\includegraphics[width=1\columnwidth]{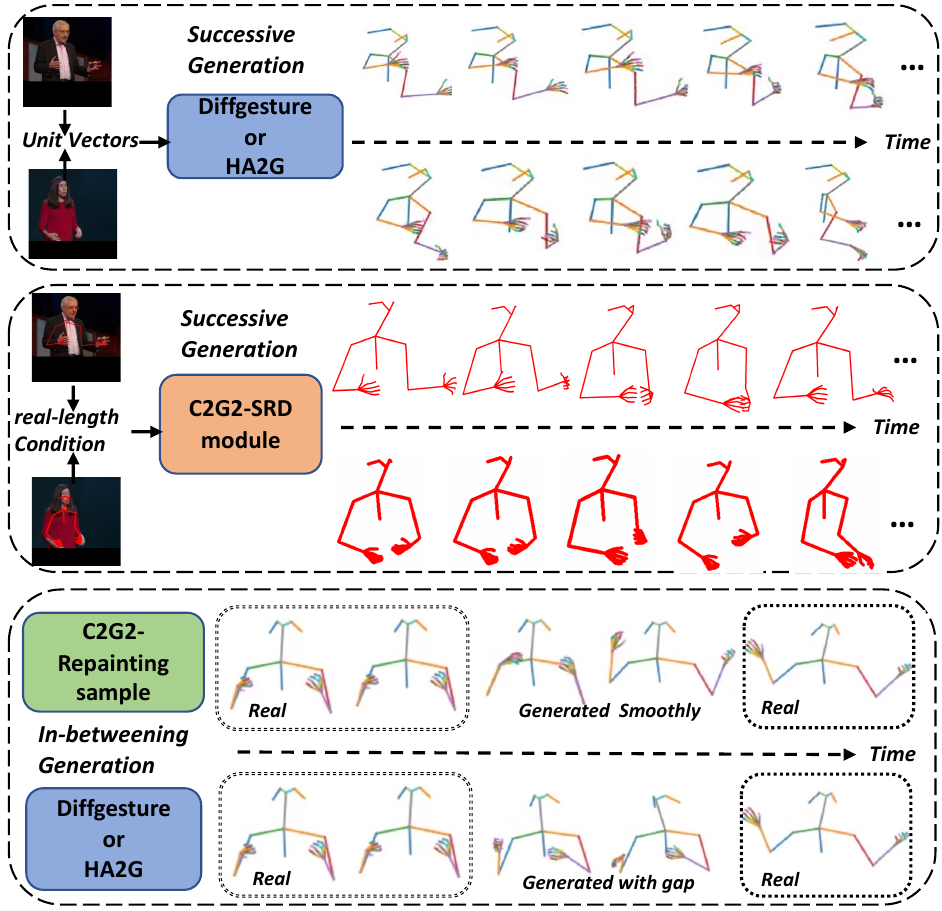} 
\caption{Controlling features of C2G2: speaker-related real-length gesture generation and flexible editing. The upper figure shows the \emph{successive} and \emph{unit-vectored} gesture generation of existing methods; the intermediate figure shows the \emph{successive} and \emph{speaker-related real-length} gesture generation by C2G2; the lower figure shows the \emph{in-betweening} gesture generation by C2G2.}
\vspace{-0.5cm}
\label{fig1}
\end{figure}

However, GAN-based approaches may encounter the issues of mode collapse and unstable training, leading to the unsatisfactory performance.
A recent work \cite{zhu2023taming} proposes a \emph{DiffGesuture} framework that uses the diffusion model \cite{ho2020denoising} for co-speech gesture generation.
By reformulating the multi-frame gesture clip as the diffusion latent space, it successfully generates clip of gestures in a non-autoregressive manner.
However, \emph{DiffGesuture} simply uses a transformer block to capture the long-term temporal dependency and only leaves the potential temporal inconsistency to be alleviated in the sampling process, which degenerates the generation quality of comprehensive finger movements.

Recently, latent diffusion model \cite{rombach2022high} is proposed for high-quality generation tasks. 
By discovering a prior latent space, the model can remove redundant information and preserve the semantic variation for the diffusion process. 
This motivates us to enhance the temporal coherence in co-speech gesture generation using a two-stage procedure. 
In the first stage, we propose to train a temporal-aware Vector Quantized Variational AutoEncoder (VQ-VAE) augmented with a cross-frame attention module, which allows the model to capture temporal dependencies among frames before the quantization step. 
This helps to learn latent temporal-aware codes for the skeleton data, which are more expressive and temporally coherent.
In the second stage, we use a diffusion process to enhance the temporal dependency between gesture and audio data, which is similar as \cite{zhu2023taming} but with a crucial difference. 
Instead of performing the diffusion process directly in the data space, we conduct the diffusion in the latent temporal-aware space obtained from the first stage. 
This allows to better model and preserve the temporal relationships between gestures and audios, leading to more synchronized and coherent co-speech gesture generation.

Moreover, we note that existing methods generate co-speech gestures with some limitations. 
On one hand, the generated results are usually skeletons represented by unit direction vectors \cite{liu2022learning,zhu2023taming} or joints from SMPL-X with consistent shape \cite{ao2023gesturediffuclip}, that generally overlook the speaker information.
On the other hand, the generation is successively done conditioned on short previous frames, which may result in uncontrollable or unreasonable gestures in a long run.
These limitations highly jeopardize the downstream tasks like human-body rendering or animation synthesis.

To this end, we propose a \emph{Controllable Co-speech Gesture Generation} framework with latent diffusion, called \textbf{C2G2}.
The C2G2 framework overcomes the limitations of previous methods by introducing two key features.
The first key feature is the inclusion of a speaker-conditioned decoder. 
By incorporating the speaker's identity or characteristics as conditioning information, C2G2 enables the generation of personalized gestures that align with the specific speaker's style and mannerisms. 
The second key feature is the introduction of a repainting strategy during the sampling process. 
This strategy provides flexibility and control over the gesture generation by allowing users to generate and edit gestures in any time intervals.
With this feature, we can generate both successive and in-betweening gestures.
Combining these two features with latent diffusion, the C2G2 framework promises to achieve more smoothed and controllable co-speech gesture generation. 
To sum up, this paper makes the following contributions:
\begin{itemize}
    \item Based on the latent diffusion model, we propose a two-stage co-speech gesture generation model with two-stage temporal dependency enhancement. Extensive experiments on benchmark datasets show the effectiveness of our proposed C2G2 compared with current state-of-the-art methods.
    \item We develop a speaker-specific control module, that is capable of generating speaker's real length skeletons instead of unit vectors.
    \item We propose a gesture editing control module through a repainting strategy to allow arbitrary gesture editing in the long-sequence generation.  
\end{itemize}

\section{Related work}
Generating synthesized gesture for corresponding text/audio has been an important research interest in fields of computer visual graphics and multimedia.
Earlier methods mainly applies rule-based searching \cite{cassell1994animated,marsella2013virtual} or statistic models \cite{yang2020statistics} to select the appropriate gesture in a pre-defined motion pool for a given speech input. 
With the development of deep learning models, more studies start to use data-driven approaches to associate speech with the corresponding gestures. 
Some works apply CNN \cite{kucherenko2019analyzing}, GRU or LSTM \cite{ishi2018speech}, Transformer \cite{bhattacharya2021text2gestures,ahuja2019language2pose} to generate gesture conditioned on texts \cite{bhattacharya2021text2gestures,ao2023gesturediffuclip}, audio \cite{li2021audio2gestures} or mixed input \cite{ahuja2020no}. 

Recent works \cite{ginosar2019learning,ferstl2020adversarial,ahuja2020style,ahuja2020no} focus on GAN based model due to its high-fidelity generation quality. 
Being a contributive work, \cite{liu2022learning} applies a hierarchical pose generator HA2G for part-aware progressive generation, and achieves natural and high-quality gestures. 
Motivated by the fact that different body parts have different levels of correlation with speech signals, Talkshow \cite{yi2023generating} leverages part-aware expert model with discrete representation to generate head, pose and hand, separately.
Following the idea of taming transformer \cite{esser2020taming}, Lu et al. \cite{lu2023co} propose to learn discrete tokens for a chunk of gestures and then auto-agressively generate theses tokens using a transformer block.
With the same idea of \cite{lu2023co}, \cite{ye2023salient} further considers the text condition in the second stage. 
Moreover, there are also methods focusing on emotive controlling by introducing style block as extra condition in diffusion or transformer generation procedure \cite{ahuja2020style,qi2023emotiongesture,ao2023gesturediffuclip}. 

Very recently, Zhu et al. \cite{zhu2023taming} propose a \emph{Diffgesture} framework that uses a diffusion model \cite{ho2020denoising} to directly generate sequential input skeletons given audio input, capable of generating high-quality gestures with detailed finger movements.
The generation results outperform previous methods.
However, although \emph{Diffgesture} applies a transformer audio-gesture noise predictor, the generated motion still has temporal inconsistency and uncontrollable collapse.
Beyond this, it still suffers from the two limitations as elaborated in the Introduction.
To overcome these issues, we propose a C2G2 framework with a two-level temporal dependency enhancement and allow speaker-related controlling and flexible editing within the latent diffusion generation.

\section{Method}
Our objective is to produce co-speech gestures that exhibit high quality, temporal smoothness, and controllability, while maintaining coherence with the input audios. 
To do so, we present a novel \emph{Controllable Co-speech Gesture Generation} framework, empowered by latent diffusion, called \textbf{C2G2}.
The overall architecture of C2G2 consists of two modules: temporal-aware VQ-VAE and gesture latent diffusion model. 
The former is designed to capture the temporal coherence of gesture sequences through a sequence of discrete latent codes. 
On the other hand, the latter employs a diffusion process to generate a diverse range of latent emebedding sequences conditioned on the input audios, encoding the temporal dependency of multi-modal inputs.
Moreover, we introduce two enhancements, a speaker-related decoder (SRD) that is conditioned on a random reference frame and a repainting strategy that empowers both successive and inbetweening gesture generation. 
The SRD decoder directly generates skeletons tied to the speaker, deviating from the conventional approach of generating unit vectors.
The repainting strategy facilitates flexible editing of generated gestures, enabling finer control over the generated animations.
The details of each module are presented as follows.

\subsection{Preliminary}
We start with the problem formulation of co-speech gesture generation.
The inputs include the upper-body gesture joint sequences and the corresponding audios.
Precisely, to obtain accurate gesture movements, state-of-the-art pose estimators Openpose \cite{cao2017realtime} and Expose \cite{choutas2020monocular} are used to extract $T$-frames joint sequences $\mathbf{x} = \{\mathbf{x}_1,\mathbf{x}_2,...,\mathbf{x}_T\}$, as well as the matched audio sequence $\mathbf{a} = \{\mathbf{a}_1,\mathbf{a}_2,...,\mathbf{a}_T\}$. 
Following \cite{yoon2020speech,liu2022learning,zhu2023taming}, a normalization step is applied to eliminate the speaker identity, resulting in the unit direction vector $\mathbf{x}_i = \{ \mathbf{d}_{i_1},...,\mathbf{d}_{i_{J-1}} \}$ with $J$ as the number of total joints and $\mathbf{d}_{i_j}$ representing the direction vector between the $j$-\emph{th} and the $(j+1)$-\emph{th} joint of the $i$-\emph{th} frame.
The reason of conducting such a normalization is for the stable training of the following models as diversified-length skeletons may easily cause model collapse.
However, to achieve the speaker-related real-length gesture generation, we also preserve real-length direction vectors $\mathbf{s}_i=\{ \mathbf{s}_{i_1},...,\mathbf{s}_{i_{J-1}} \}$. 
In C2G2, a temporal-aware VQ-VAE is first trained to reconstruct the unit direction vectors $\mathbf{x}_i$s, and then a speaker-related decoder is trained to generate the real-length direction vectors $\mathbf{s}_i$s conditioned on arbitrary reference frame $\mathbf{s}_r$.
The reverse process of the latent diffusion model learns to synthesize the latent embedding $\{\mathbf{z}_1,\mathbf{z}_2,...,\mathbf{z}_T\}$ given the audio input $\mathbf{a}$ and the latent embedding of $N$ previous frames, $\{\mathbf{z}_1,\mathbf{z}_2,...,\mathbf{z}_N\}$, which is obtained from VQ-VAE.

\subsection{Temporal-aware VQ-VAE}
We first introduce a temporal-aware VQ-VAE to learn the latent codes for input joints sequences.
Given a gesture sequence $\mathbf{x} = [\mathbf{x}_1,\mathbf{x}_2,...,\mathbf{x}_T]$, VQ-VAE aims to reconstruct the gesture sequence through a temporal-aware encoder-decoder structure and a quantized codebook \emph{C} containing $k$ codes . 
The fundamental architecture of the proposed VQ-VAE is a symmetric encoder-decoder structure as shown in the top half of Figure \ref{vqvae}. 
To better quantize comprehensive sequence containing the fingers of both hands, the encoder \emph{E} consists of two stacking modules: a content extractor that is composed with linear based ResNet blocks, and a context encoder that uses cross-frame attention modules with positional encoding to capture the temporal coherence embedded in the long-term sequence.
It maps the input sequence to a latent embedding $\mathbf{z}=[\mathbf{z}_1,\mathbf{z}_2,...,\mathbf{z}_T]$.
Afterwards, the quantizer \emph{Q} finds the corresponding latent codes $\mathbf{c}=[\mathbf{c}_1,\mathbf{c}_2,...,\mathbf{c}_T]$ from the codebook \emph{C} by optimizing:
$$\mathbf{c}_i = arg min||\mathbf{z}_i-\mathbf{c}_i||_2, \mathbf{c}_i \in C.$$
The matched latent codes are then feed into the decoder \emph{D} to reconstruct the sequence. 
The overall learning objective contains two components: a reconstruction loss and a commitment loss, where the former effectively measures the generating quality and the latter encloses the distance between the codes and latent features.
\begin{figure}[t]
\centering
\resizebox{\linewidth}{!}{
\includegraphics[width=1.4\columnwidth]{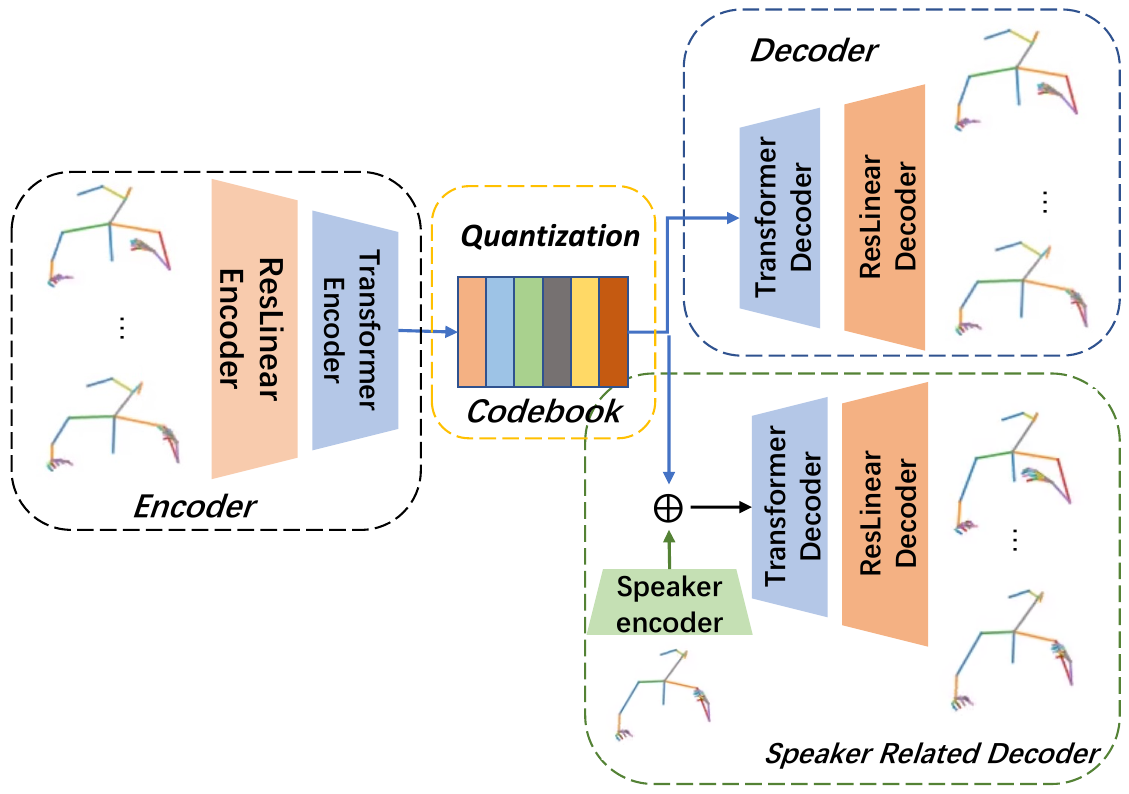} }
\caption{The structure of the temporal-aware VQVAE.}
\label{vqvae}
\vspace{-0.5cm}
\end{figure}


\subsubsection{EMA Quantization Strategy}
In the training, the quantizer may easily stuck into the codebook collapse issue due to the high dimensionality of the training data. 
To provide smooth codebook updating, exponential moving average (EMA) and codebook reset strategies are implemented to avoid possible collapse due to inactive codes \cite{zhang2023generating,razavi2019generating}. 
Reset strategy replaces the inactive codes according to the output of the encoder. 
EMA smooths the upgrading of the codebook \emph{C} by $C_T = \mu C_{T-1}+(1-\mu)C_T$. 
We provide an ablation study for the quantizer structure in experimental studies.
\begin{figure*}[t]
\centering
\resizebox{\linewidth}{!}{
\includegraphics[width=2\columnwidth]{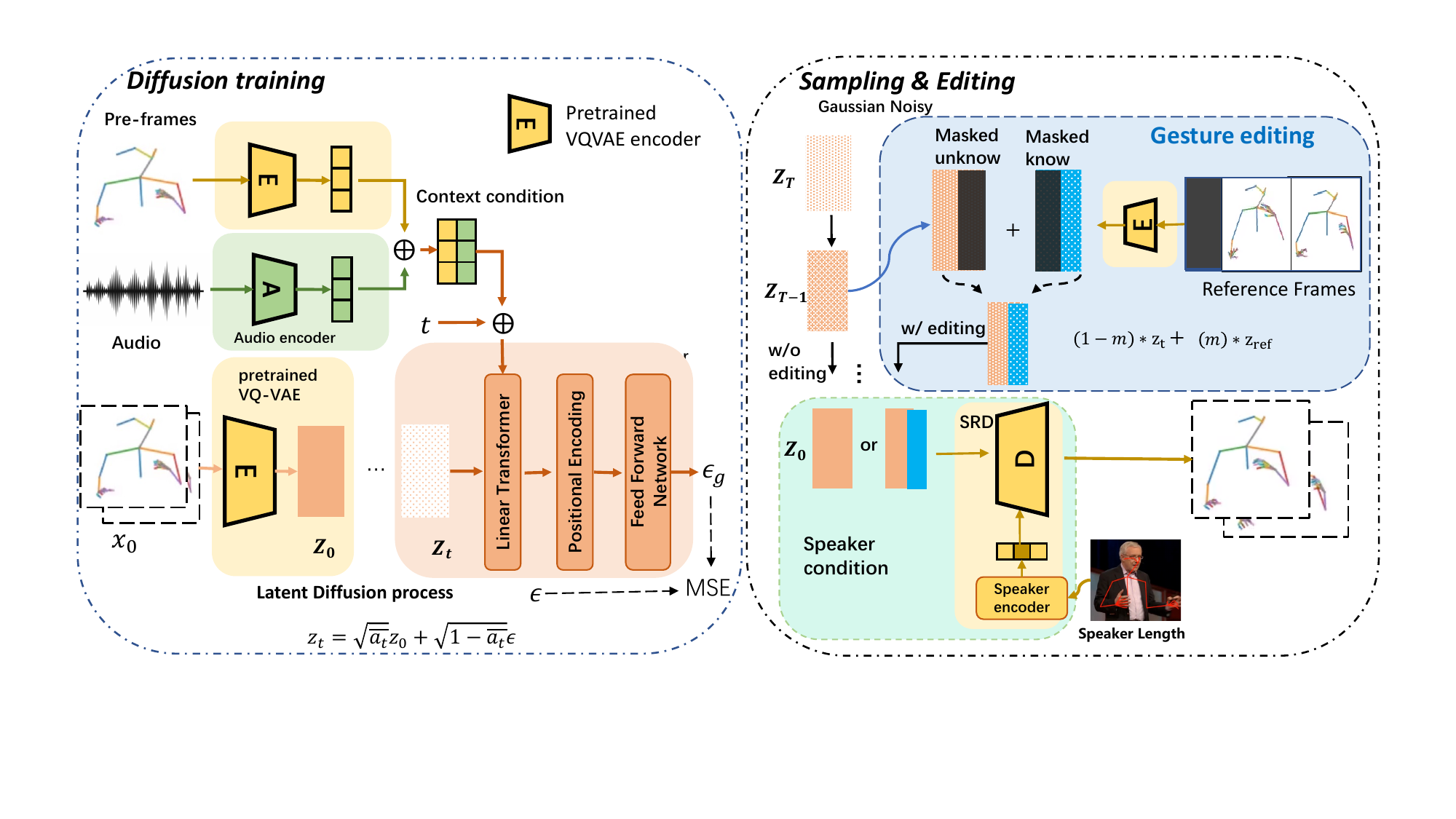}} 
\caption{The framework overview of C2G2. Given input skeleton sequence $\mathbf{x}$, a pre-trained VQ-VAE will be used to generate latent embedding $\mathbf{z}_0$ in latent space representation. Then diffusion forward process will turn $\mathbf{z}_0$ into $\mathbf{z}_T$. A transformer based noise predictor will be applied for reverse denoising procedure. A speaker related real-length decoder will enable speaker controlling by extracting length information from given speaker condition. During sampling, we propose a repainting sampling strategy for in-betweening generation. }
\label{fig1}
\vspace{-0.5cm}
\end{figure*}

\subsubsection{Speaker Related Decoder}
The above structure works on the input sequence represented by unit direction vectors and thus overlooks the speaker identity information. 
To generate the speaker specific real-length gestures, we further propose a speaker-conditioned decoder $D_s$ as shown in the bottom half of Figure \ref{vqvae}.
The main difference of $D_s$ from \emph{D} is on an extra speaker encoder.
The speaker encoder, which is composed of MLP and convolutional layers, is used to extract the speaker identity information from an arbitrary prompting frame $\mathbf{s}_r$, and then concatenate it with the latent codes.
Note that, to keep the previously learned quantized code space stable, the encoder \emph{E} and the quatizer \emph{Q} are frozen and only the speaker related decoder is trained on the real-length direction vectors. 

\subsection{Latent Diffusion Model}
To generate reliable and diverse gestures, we leverage latent diffusion model (LDM) for the latent embedding generation.
The core idea of LDM is to gradually denoise the Gaussian noise \cite{dhariwal2021diffusion,rombach2022high,huang2023make}. 
Formally, LDM defines bi-directional process: forward diffusion and reverse denoising.

In forward process, LDM assumes that the process of adding noise based on the variance schedule $\beta_t$ follows the Markov chain \cite{rombach2022high}.
Through variances schedule $\beta_t$, the input distribution is finally corrupted into a pure noise with the Gaussian distribution $\mathbb{N}(\mathbf{z}_T;\mathbf{0},\mathbf{I})$. 
Based on the independent property of the noise schedule, given $\bar\alpha_t=\prod_{s=1}^t(1-\beta_s)$, the forward process can be summarized into one step:
\begin{equation} \label{diffeq3}
q(\mathbf{z}_t|\mathbf{z}_0):=\mathbb{N}(\mathbf{z}_t;\sqrt{\bar\alpha_t}\mathbf{z}_0,(1-\bar\alpha_t)\mathbf{I}).
\end{equation}

Reverse denoising, i.e., the generation process, models the conditional distribution of $p_\theta(\mathbf{z}_{t-1}|\mathbf{z}_t,\boldsymbol{\Phi})$ through the predicted noise given the context $\boldsymbol{\Phi}$. 
For the co-gesture generation task, $\boldsymbol{\Phi}$ is the combination of the latent embedding of previous \emph{N} frames $\{ \mathbf{z}_1,.., \mathbf{z}_N\}$ and audio features $\mathbf{a}$. 
\begin{equation} \label{diffeq4}
p_\theta(\mathbf{z}_{t-1}|\mathbf{z}_t,\boldsymbol{\Phi})=\mathbb{N}(\mathbf{z}_{t-1};\mu_\theta(\mathbf{z}_t,t,\boldsymbol{\Phi}),\beta_t \mathbf{I}).
\end{equation}

To further boost the temporal dependency across frames for the latent sequence, we leverage the context measuring capacity of transformer for sequence modelling. 
Contextual information including previous \emph{N} frames and audio features are concatenated together with $\mathbf{z}_t$ in the feature channel. 
A self-attention module is then applied to discover the long-term temporal dependency.
In the sampling process, to achieve the trade-off between diversity and temporal consistency, we adopt the \emph{thresholding}, \emph{smoothing sampling} and \emph{classifier-free guidance} tricks as used in \cite{zhu2023taming}.

\subsection{In-betweening Gesture Generation}
To achieve flexible co-speech gesture generation, we further consider the in-betweening generation scenario where the first $N$ frames and the last $M$ frames are given and the aim is to generate intermediate temporally smooth gestures.
Existing methods only support for successive generation, and generally fail in this scenario.
\begin{table*}[t]
\centering
\renewcommand{\arraystretch}{1.3}
\begin{tabular}{ccccccc}
\hline
\multicolumn{1}{c}{\multirow{2}{*}{\textbf{Methods}}} & \multicolumn{3}{c}{\textbf{TED Gesture} }                                                   & \multicolumn{3}{c}{\textbf{TED Expressive}}                                                 \\ \cline{2-7}
 &FGD $\downarrow$ &BC $\uparrow$&Diversity $\uparrow$ &FGD $\downarrow$ &BC $\uparrow$&Diversity $\uparrow$\\
\hline
Ground Truth                                            & 0.000                    & 0.698                   & 108.525                        & 0.000                    & 0.703                   & 178.827                        \\ 
VQ-VAE Reconstruction                                            & 0.109                    & 0.636                   & 107.450                        & 0.402                    & 0.703                   & 179.743                        \\
\hline    
Att-S2S \cite{yoon2019robots}                                           & 18.154                   & 0.196                   & 82.776                         & 54.920                   & 0.152                   & 122.693                        \\
S2G \cite{ginosar2019gestures}                                          & 19.254                   & 0.668                   & 93.802                         & 54.650                   & 0.679                   & 142.489                        \\
L2P  \cite{ahuja2019language2pose}                                         & 22.083                   & 0.200                   & 90.138                         & 64.555                   & 0.130                   & 120.627                        \\
TriM  \cite{yoon2020speech}                                              & 3.729                    & 0.667                   & 101.247                        & 12.613                   & 0.563                   & 154.088                        \\
HA2G \cite{liu2022learning}                                                   & 3.072                    & \textit{\textbf{0.672}} & 104.322                        & 5.306                    & 0.641                   & 173.899                        \\
DiffGesutre   \cite{zhu2023taming}                                          & 1.506                    & \textbf{0.699}          & \textit{\textbf{106.722}}      & 2.600                    & \textbf{0.718}          & 182.757           \\ \hline
C2G2 (Ours)                                                   & \textit{\textbf{1.035}}  & 0.647                   & \textbf{106.991}               & \textit{\textbf{1.168}}  & 0.708                   & \textit{\textbf{182.824}}      \\
C2G2-\emph{rl} (Ours)                                                & \textbf{0.652}           & 0.615                   & 73.066                         & \textbf{1.095}           & \textit{\textbf{0.716}} & \textbf{192.716}      \\ \hline   
\end{tabular}
\caption{The quantitative results on TED Gesture and TED expressive datasets. We compare our C2G2 against ground-truth target and recent state-of-the-art methods. Herein, \emph{rl} is to mark the model with SRD to generate speaker conditioned real-length gestures. The winner of each metric is highlighted using bold, and the runner-up is highlighted using bold and italic.}
\label{cpresults}
\vspace{-0.3cm}
\end{table*}

In our C2G2, we propose a repainting strategy, motivated by \cite{lugmayr2022repaint}, to generate the intermediate sequences by sampling from a mask region of conditional input.
For every reverse step, with the given gesture frames $\mathbf{Z}_c = [\mathbf{Z}_{f}, \mathbf{Z}_{l}]$, where $\mathbf{Z}_f$ and $\mathbf{Z}_l$ represent the latent embedding of the first and the last frames, respectively, we can sample $\mathbf{z}_{t-1}$ using:
\begin{equation} \label{repainteq1}
\mathbf{z}_{t-1}^{known}\sim \mathbb{N}(\sqrt{\bar\alpha_t}\mathbf{Z}_c,(1-\bar\alpha_t)\mathbf{I}),
\end{equation}
\begin{equation} 
\mathbf{z}_{t-1}^{unknown}\sim \mathbb{N}(\mu_\theta(\mathbf{z}_t,t,\boldsymbol{\Phi}),\beta_t \mathbf{I}),
\end{equation}
\begin{equation}
\mathbf{z}_{t-1} =\mathbf{m}\cdot \mathbf{z}_{t-1}^{known}+(1-\mathbf{m})\cdot \mathbf{z}_{t-1}^{unknown},
\end{equation}
where $\mathbf{m}$ is a mask indicating the unknown and known areas.
Note that this repainting strategy does not require the re-training of LDM, and only works in the sampling process. 
With this strategy, we allow in-betweening editing in a long-term gesture sequence generation. 
Especially during auto-regressive generation process, considering the randomness brought by probabilistic modelling, such flexible editing can be used in any step to interpolate expected gestures like hand shaking for reliable controlling.

\section{Experimental Studies}
In this section, we first compare C2G2 with several state-of-the-art co-speech gesture generation baselines, and then conduct comprehensive property analyses of C2G2.

\subsection{Datasets Description}
\textbf{Ted Gesture} \cite{yoon2020speech} is a large-scale English-based dataset for co-speech gesture generation, composed of 1766 TED videos from various topics and different narrators.
The 3D gesture joints of human upper body as well as the corresponding audio sequence are available. 
Following the data pre-processing of previous works \cite{zhu2023taming,razavi2019generating,yoon2020speech}, pose frames are sampled in 15 FPS
A clip segmentation consists of 34 video frames with 10-frame as the stride. 
Both unit direction and real-length vectors are extracted for gesture representation.

\noindent \textbf{Ted Expressive}  \cite{liu2022learning} is a recent English-based dataset for co-speech gesture generation. 
Different from Ted Gesture dataset that only contains body skeletons, Ted Expressive extracts 30 finger joints using ExPose \cite{ExPose:2020} for each frame to harvest diverse and detailed gesture representation.
Based on SMPL-X \cite{SMPL-X:2019} 3D human parameterized model, ExPose joints can be directly turned into shape related human poses for further rendering. 
Same as in Ted Gesuture, both unit direction and real-length vectors are preserved for the following gesture generation.

\subsection{Experimental Settings}
\textbf{Comparing baselines}. We compare our C2G2 with: 1) \textbf{Attention Seq2Seq} (Att-S2S) \cite{yoon2019robots} that leverages attention mechanisms for temporal gesture generation given texts; 2) \textbf{Speech2Gesture} (S2G) \cite{ginosar2019gestures} that uses cross-model translation model taken audio as input for movement generation; 3)
\textbf{Language2Pose} (L2P) \cite{ahuja2019language2pose} that learns a joint embedding space for language and pose clip; 4) \textbf{Trimodel} (TriM) \cite{yoon2020speech} that combines information from text, audio and speaker identity;
5) \textbf{HA2G} \cite{liu2022learning} that applies hierarchical audio-gesture generator across multiple level semantic granularity; and 6) \textbf{Diffgesture} \cite{zhu2023taming} that uses diffusion based model with sampling stablizer for diverse gesture generation. 

\noindent \textbf{Implementation Details.} 
For all the experiments, we sample $34$ frames for generation and $4$ precondition reference frames following \cite{yoon2020speech}. 
The number of joints is $10$ and $43$ for TED Gesture and TED expressive, respectively. 
For the temporal-aware VQVAE, the content extractor is composed of 8 layers of linear Residual blocks, and the temporal extractor is a $R$-layer transformer encoder containing positional embedding, self attention and feed-forward Network where $R=4$ in the encoder and $R=2$ in the decoder. 
The quantizer is based on EMAReset as mentioned in Section 3.1 where the updating rate $\mu=0.9$. 
For SRD, the speaker reference encoder is composed of 3 layers of Residual blocks that outputs a 64-dimensional embedding vector. 
For the latent codebook with $N_C$ $d_{C}$-dimensional codes, $N_{C}=1024,d_{C}=64$ for TED Gesture and $N_{C}=1024,d_{C}=128$ for TED Expressive. 
For the audio encoder, we use the structure in \cite{yoon2020speech} to extract information from raw audios.
A conditional frame encoder that is used to process the given 4 frames share the weights of VQ-VAE encoder.

For the latent diffusion module, the denoising step is $500$ and the variance schedule is linearly increasing from $1e-4$ to $0.02$. 
The transformer module is a 8-layers transformer encoder with self-attention and feed-forward Network. 
The hidden dimension of the transformer is 256 for both TED Gesture and TED Expressive. 
We use Adam as the optimizer with the learning rate $5e-4$ for TED Gesture and $3e-4$ for TED Expressive. 
The training is done on a single NVIDIA A100 GPU.
For the VQ-VAE training, it tasks $\sim$6 hours for TED Gesture and $\sim$14 hours on TED Expressive.
For the diffusion module training, it takes $\sim$8 hours on TED Gesture and $\sim$20 hours on TED Expressive.
\begin{figure*}[t]
\centering
\includegraphics[width=2\columnwidth]{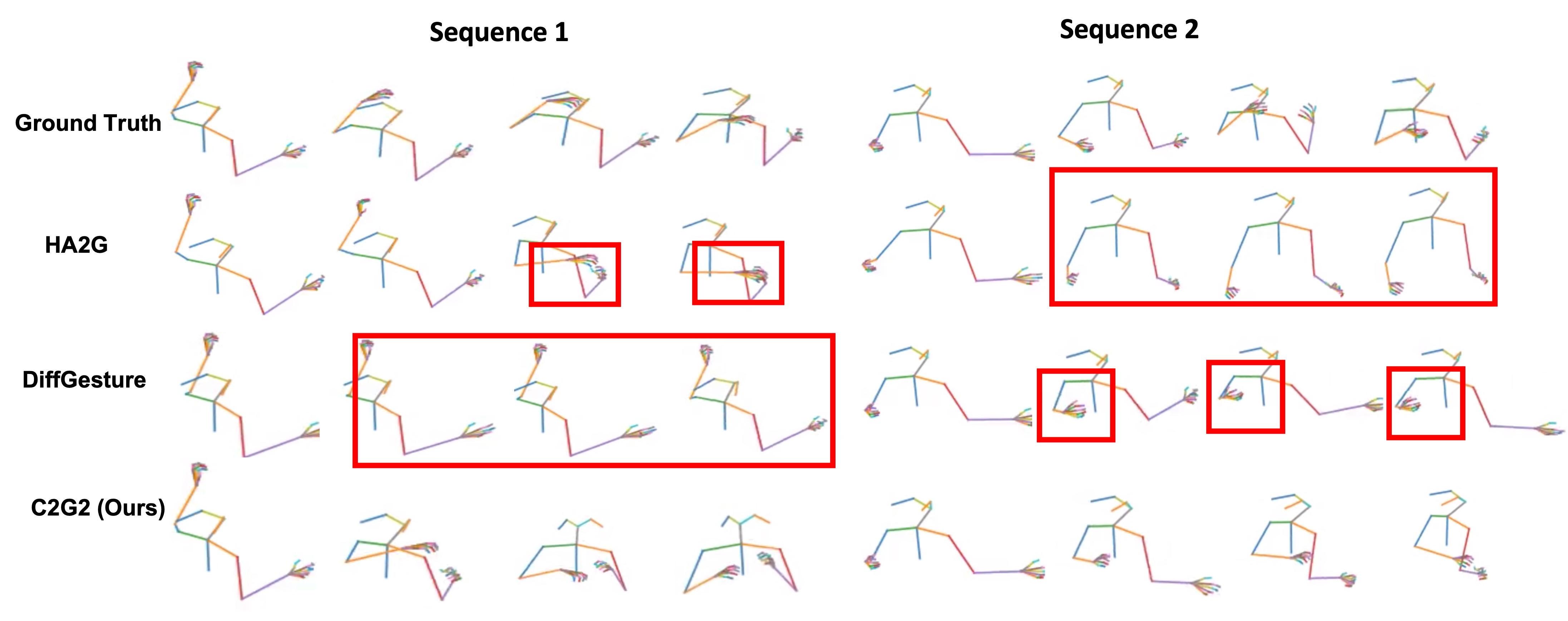} 
\caption{The visualization results of generated gesture sequence from C2G2 and previous methods \cite{liu2022learning,zhu2023taming}. 
Two sequences, each with four continuous frames, are selected. We highlight the slowly changed or nearly frozen frames in the generated results with red rectangles.  }
\label{visualization}
\vspace{-0.2cm}
\end{figure*}

\noindent \textbf{Evaluation Metrics.}
We adopt three widely used metrics to measure the naturalness and correctness of the generated gesture movements, namely Frechet Gesture Distance (FGD), Beat Consistency Score (BC), and Diversity.
More description on these metrics can be found in the Appendix.

\subsection{Evaluation Results}
\textbf{Quantitative Results.} 
We first compare C2G2 with existing baselines on the above 3 metrics. 
For the real-length skeleton generation obtained from C2G2-\emph{rl}, we use the same FGD model in HA2G \cite{liu2022learning} but normalize the final results back to unit direction vectors for the fair comparision. 
All the comparison results are shown in Table \ref{cpresults}.
From the table, we observe that VQ-VAE reconstruction achieves comparable results with the ground-truth ones, which shows the excellent reconstruction capability of the proposed temporal-aware VQ-VAE.

For the metric FGD, both C2G2 and C2G2-\emph{rl} achieve significant improvements on the two datasets.
Specifically,C2G2-\emph{rl} is consistently the winner among all the baselines, achieving 0.854 and 1.505 lower values than the best baseline DiffGesture.
C2G2 is the runner-up, which yields better results than all existing baselines but inferior results than C2G2-\emph{rl}.
These results demonstrate that our C2G2 can generate more realistic gestures as FGD mainly reflects the distance of the generated distribution with the real one.

For the metric diversity, although C2G2-\emph{rl} obtains the best result on TED Expressive, we also note that it fails on TED Gesture.
We suspect the reason is because the gesture space of TED Gesture is not that diversified and C2G2-\emph{rl} is overfitted, as indicated by its smallest FGD value, and thus constrains the generation diversity. 
However, such an inferiority disappears on TED Expressive containing more complicated figure movements. 
As can be seen, C2G2-\emph{rl} is a clear winner among all the baselines for diversity on TED Expressive.

Finally, for the metric BC, C2G2-\emph{rl} and C2G2 generally achieve competitive results with the best baseline DiffGesture.
Future studies will put the focus on more powerful audio encoder to further boost the BC performance. 

\noindent \textbf{Qualitative Results.}
In this section, we visualize the generated gestures from different methods for the qualitative comparison.
Considering that HA2G \cite{liu2022learning} and DiffGesture \cite{zhu2023taming} outperform the other baselines as presented in their paper, we mainly compare C2G2 with these two methods.
The visualization results are illustrated in Figure \ref{visualization}. 
As can be seen, HA2G generates wired and slow gesture movements without proper matching with the input audio.
This is because HA2G may suffer from the model collapse issue during training.
Diffgesture generally achieves good continuity, however, it is easily stuck into one gesture for several frames, leading to a weak correlation with the input audios. 
Among the three methods, C2G2 can ensure both the temporal smoothness and semantic correlation with the input audios.
\begin{table}[t]
\centering
\renewcommand{\arraystretch}{1.3}
\resizebox{\linewidth}{!}{
\begin{tabular}{cccccc}
\hline
\multicolumn{2}{c}{Metric}                 & Ground-truth   & HA2G  & Diffgesture & C2G2           \\ \hline
\multirow{3}{*}{Short Clips}  & NAL & 3.438          & 3.275 & 3.144       & \textbf{3.506} \\
                              & SMTH & 3.325          & 3.144 & 3.144       & \textbf{3.681} \\
                              & SYN  & \textbf{3.494} & 3.331 & 3.294       & 3.350          \\ \hline \hline
\multirow{3}{*}{Long Clips}   & NAL & 2.916          & 3.056 & 2.741       & \textbf{3.394} \\
                              & SMTH & 2.894          & 2.992 & 2.870       & \textbf{3.384} \\
                              & SYN  & \textbf{3.411} & 3.028 & 3.247       & 3.304          \\ \hline \hline
\multirow{3}{*}{Middle Clips} & NAL & \textbf{4.009} & 3.275 & 3.456       & 3.903          \\
                              & SMTH & 3.894          & 2.759 & 2.850       & \textbf{4.072} \\
                              & SYN  & \textbf{4.128} & 3.175 & 3.734       & 3.881          \\ \hline \hline
\multirow{3}{*}{Average}      & NAL & 3.469          & 3.219 & 3.163       & \textbf{3.619} \\
                              & SMTH & 3.363          & 3.003 & 3.025       & \textbf{3.738} \\
                              & SYN  & \textbf{3.647} & 3.216 & 3.413       & 3.522     \\ \hline    
\end{tabular}}
\caption{User study. MOS scores for the naturalness (NAL), smoothness (SMTH) and synchrony (SYN) of the generated gesture sequences. Short and long clips are for successive generation, and middle clips are for in-betweening generation. We compare C2G2 with HA2G and Diffgesture.}
\label{MOS}
\vspace{-0.6cm}
\end{table}

\noindent \textbf{User Study.} 
We further conduct a subjective user study to measure the naturalness, smoothness and synchrony of the generated gesture sequences for different methods. 
Specifically, we randomly sample 20 clips from TED Expressive, including 10 short clips (each with $\sim$2s), 5 long clips (each with $\sim$60s) and 5 middle-length clips (each with $\sim$10s).
The short and long clips are used to evaluate the short-term and long-term successive generation quality, while the middle-length clips are used for the evaluation of the in-betweening generation.
Specifically, for each middle-length clip, the first 4 frames and the last $\sim$25 frames are pre-given, and the intermediate frames are to be generated.
We invite 8 volunteers to give the score for the generated gesture sequences from 3 perspectives: naturalness, smoothness and synchrony.
The ratings are from 1 to 5 with 5 marking the highest quality. 
The detailed scoring rules are presented in the Appendix.
Mean Opinion Score (MOS) is then calculated based on the evaluation results and is shown in Table \ref{MOS}.

From Table \ref{MOS}, we can see that C2G2 consistently ensures the better naturalness, synchrony and smoothness compared with Diffgesture and HA2G in average. 
More specifically, all the methods generally perform better in short clips than in long clips. 
This is reasonable as long sequences indicate more complex and diversified gestures included.
It is surprising that our generated gestures have even higher naturalness and smoothness than the Ground-Truth ones, which demonstrates the superiority of C2G2 in the successive generation scenario. 
Moreover, C2G2 achieves more significant improvements in middle clips generation, where the repainting strategy is applied, especially on smoothness. 
This verifies that C2G2 can also generate high-quality gestures in the in-betweening generation scenario, while existing methods usually have an obvious gap in the transition part between the generated and real frames.
All these results prove that C2G2 is a promising co-speech gesture generation model.
\begin{table}[t]
\centering
\renewcommand{\arraystretch}{1.3}
\resizebox{\linewidth}{!}{
\begin{tabular}{ccccc}
\hline
\multicolumn{1}{c}{\multirow{2}{*}{\textbf{Methods}}} & \multicolumn{2}{c}{Ted Gesture} & \multicolumn{2}{c}{Ted Expressive} \\ \cline{2-5}
\multicolumn{1}{c}{}                                  & FGD$\downarrow$           & Diversity $\uparrow$       & FGD$\downarrow$            & Diversity$\uparrow$         \\ \hline
Ground Truth                                          & 0.000         & 108.525         & 0.000          & 178.827           \\ \hline
V$_{w/o\_enc\_trans}$                                        & 3.959         & 98.707          & 4.939          & 170.039           \\
V$_{w/o\_dec\_trans}$                                        & 10.098        & 95.335          & 23.292         & 158.320           \\
Full VQ-VAE                                           & \textbf{0.109}         & \textbf{107.450}         & \textbf{0.402 }         & \textbf{179.743}          \\ \hline
\end{tabular}
}
\caption{Ablation study results of encoder-decoder structure w/o temporal transformer module.}
\label{ab1}
\vspace{-0.2cm}
\end{table}

\subsection{Ablation Study}
\noindent \textbf{Ablation Study on VQ-VAE Structure.} 
To verify the temporal coherence enhancement brought by the temporal-aware VQ-VAE, we propose to compare with two more variants.
The first variant is a VQ-VAE without transformer module in the encoder, and the second one is a VQ-VAE without transformer module in the decoder. 
We compare the two variants with the full VQ-VAE with respect to FGD and diversity on both TED Gesture and Expressive datasets.  
The comparison results are shown in Table \ref{ab1}.
From Table \ref{ab1}, we can easily observe that transformer layers are both necessary in encoder and decoder to improve the quantization performance by capturing temporal information.

\noindent \textbf{Ablation Study on Quantizer of VQ-VAE.}
Compared with TED Gesture, TED Expressive contains more complicated information due to fine-grained finger keypoints (8 basic skeleton keypoints and 15 keypoints per hand). 
Thus, a more powerful quantizer is crucial for the quantization of the gesture sequences in TED Expressive. 
We conduct ablation studies with respect to EMA and Reset strategies on TED Expressive.
We mainly compare with the quantizer (1) without both EMA and Reset strategy, (2) without EMA strategy, and (3) without Reset Strategy.
The results are shown in Tabel \ref{ab2}.
As can be seen, quantizer with both EMA codebook updating and reset of inactive codes can achieve good reconstruction performance with the lowest FGD and competitive diversity.
\begin{table}
\centering
\renewcommand{\arraystretch}{1.1}
\resizebox{\linewidth}{!}{
\begin{tabular}{ccccc} 
\toprule 
Method &EMA&Reset& FGD $\downarrow$ & Diversity $\uparrow$\\
\hline
Ground Truth&&&0&178.827\\
\hline
Quantizer &&&21.495&151.653\\
Quantizer &\checkmark&&0.670&180.829\\
Quantizer&&\checkmark&0.522&\textbf{180.876}\\
Quantizer &\checkmark&\checkmark&\textbf{0.402}&{179.743}\\
\bottomrule 
\end{tabular}}
\caption{Ablation study results of different quantizer w/o EMA or Reset strategies on TED Expressive.}
\label{ab2}
\vspace{-0.2cm}
\end{table}

\section{Conclusion}
In this paper, we propose a novel latent diffusion based co-speech generation framework named \textbf{C2G2} that enables two different gesture controlling. 
To better enhance the temporal consistency, we leverage a two-stage latent diffusion model composed of a temporal-aware VQ-VAE and transformer based diffusion noise predictor.
Moreover, considering more practical scenarios, we allow speaker's real-length conditioning and movement editing through SRD module and repainting sampling strategy. 
Extensive experimental studies including both quantitative and qualitative evaluations demonstrate the effectiveness of our C2G2 compared with several state-of-the-art baselines.
We also remark some limitations and future working directions for our work.
Firstly, the pre-trained models may contain bias towards a single language as it is trained on an English-based dataset. Multi-linguistic dataset collection is a worthy direction for the community for more general co-speech gesture generation. 
Moreover, our model composes of two-stage training as well as one more step of speaker decoder finetuning, thus may result in unstable generation quality due to the stage-level error propagation.

\bibliography{aaai24}

\begin{thebibliography}{50}
\providecommand{\natexlab}[1]{#1}

\bibitem[{Ahuja et~al.(2020{\natexlab{a}})Ahuja, Lee, Ishii, and
  Morency}]{ahuja2020no}
Ahuja, C.; Lee, D.~W.; Ishii, R.; and Morency, L.-P. 2020{\natexlab{a}}.
\newblock No gestures left behind: Learning relationships between spoken
  language and freeform gestures.
\newblock In \emph{Findings of the Association for Computational Linguistics:
  EMNLP 2020}, 1884--1895.

\bibitem[{Ahuja et~al.(2020{\natexlab{b}})Ahuja, Lee, Nakano, and
  Morency}]{ahuja2020style}
Ahuja, C.; Lee, D.~W.; Nakano, Y.~I.; and Morency, L.-P. 2020{\natexlab{b}}.
\newblock Style transfer for co-speech gesture animation: A multi-speaker
  conditional-mixture approach.
\newblock In \emph{Computer Vision--ECCV 2020: 16th European Conference,
  Glasgow, UK, August 23--28, 2020, Proceedings, Part XVIII 16}, 248--265.
  Springer.

\bibitem[{Ahuja and Morency(2019)}]{ahuja2019language2pose}
Ahuja, C.; and Morency, L.-P. 2019.
\newblock Language2pose: Natural language grounded pose forecasting.
\newblock In \emph{2019 International Conference on 3D Vision (3DV)}, 719--728.
  IEEE.

\bibitem[{Alexanderson et~al.(2020)Alexanderson, Henter, Kucherenko, and
  Beskow}]{alexanderson2020style}
Alexanderson, S.; Henter, G.~E.; Kucherenko, T.; and Beskow, J. 2020.
\newblock Style-controllable speech-driven gesture synthesis using normalising
  flows.
\newblock In \emph{Computer Graphics Forum}, volume~39, 487--496. Wiley Online
  Library.

\bibitem[{Ao et~al.(2022)Ao, Gao, Lou, Chen, and Liu}]{ao2022rhythmic}
Ao, T.; Gao, Q.; Lou, Y.; Chen, B.; and Liu, L. 2022.
\newblock Rhythmic gesticulator: Rhythm-aware co-speech gesture synthesis with
  hierarchical neural embeddings.
\newblock \emph{ACM Transactions on Graphics (TOG)}, 41(6): 1--19.

\bibitem[{Ao, Zhang, and Liu(2023)}]{ao2023gesturediffuclip}
Ao, T.; Zhang, Z.; and Liu, L. 2023.
\newblock GestureDiffuCLIP: Gesture diffusion model with CLIP latents.
\newblock \emph{arXiv preprint arXiv:2303.14613}.

\bibitem[{Bhattacharya et~al.(2021)Bhattacharya, Rewkowski, Banerjee, Guhan,
  Bera, and Manocha}]{bhattacharya2021text2gestures}
Bhattacharya, U.; Rewkowski, N.; Banerjee, A.; Guhan, P.; Bera, A.; and
  Manocha, D. 2021.
\newblock Text2gestures: A transformer-based network for generating emotive
  body gestures for virtual agents.
\newblock In \emph{2021 IEEE virtual reality and 3D user interfaces (VR)},
  1--10. IEEE.

\bibitem[{Cao et~al.(2017)Cao, Simon, Wei, and Sheikh}]{cao2017realtime}
Cao, Z.; Simon, T.; Wei, S.-E.; and Sheikh, Y. 2017.
\newblock Realtime multi-person 2d pose estimation using part affinity fields.
\newblock In \emph{Proceedings of the IEEE conference on computer vision and
  pattern recognition}, 7291--7299.

\bibitem[{Cassell et~al.(1994)Cassell, Pelachaud, Badler, Steedman, Achorn,
  Becket, Douville, Prevost, and Stone}]{cassell1994animated}
Cassell, J.; Pelachaud, C.; Badler, N.; Steedman, M.; Achorn, B.; Becket, T.;
  Douville, B.; Prevost, S.; and Stone, M. 1994.
\newblock Animated conversation: rule-based generation of facial expression,
  gesture \& spoken intonation for multiple conversational agents.
\newblock In \emph{Proceedings of the 21st annual conference on Computer
  graphics and interactive techniques}, 413--420.

\bibitem[{Choutas et~al.(2020{\natexlab{a}})Choutas, Pavlakos, Bolkart,
  Tzionas, and Black}]{ExPose:2020}
Choutas, V.; Pavlakos, G.; Bolkart, T.; Tzionas, D.; and Black, M.~J.
  2020{\natexlab{a}}.
\newblock Monocular Expressive Body Regression through Body-Driven Attention.
\newblock In \emph{European Conference on Computer Vision (ECCV)}, 20--40.

\bibitem[{Choutas et~al.(2020{\natexlab{b}})Choutas, Pavlakos, Bolkart,
  Tzionas, and Black}]{choutas2020monocular}
Choutas, V.; Pavlakos, G.; Bolkart, T.; Tzionas, D.; and Black, M.~J.
  2020{\natexlab{b}}.
\newblock Monocular expressive body regression through body-driven attention.
\newblock In \emph{Computer Vision--ECCV 2020: 16th European Conference,
  Glasgow, UK, August 23--28, 2020, Proceedings, Part X 16}, 20--40. Springer.

\bibitem[{Dhariwal and Nichol(2021)}]{dhariwal2021diffusion}
Dhariwal, P.; and Nichol, A. 2021.
\newblock Diffusion models beat gans on image synthesis.
\newblock \emph{Advances in neural information processing systems}, 34:
  8780--8794.

\bibitem[{Ellis(2007)}]{ellis2007beat}
Ellis, D.~P. 2007.
\newblock Beat tracking by dynamic programming.
\newblock \emph{Journal of New Music Research}, 36(1): 51--60.

\bibitem[{Esser, Rombach, and Ommer(2020)}]{esser2020taming}
Esser, P.; Rombach, R.; and Ommer, B. 2020.
\newblock Taming Transformers for High-Resolution Image Synthesis.
\newblock arXiv:2012.09841.

\bibitem[{Ferstl, Neff, and McDonnell(2020)}]{ferstl2020adversarial}
Ferstl, Y.; Neff, M.; and McDonnell, R. 2020.
\newblock Adversarial gesture generation with realistic gesture phasing.
\newblock \emph{Computers \& Graphics}, 89: 117--130.

\bibitem[{Ginosar et~al.(2019{\natexlab{a}})Ginosar, Bar, Kohavi, Chan, Owens,
  and Malik}]{ginosar2019gestures}
Ginosar, S.; Bar, A.; Kohavi, G.; Chan, C.; Owens, A.; and Malik, J.
  2019{\natexlab{a}}.
\newblock Learning Individual Styles of Conversational Gesture.
\newblock In \emph{Computer Vision and Pattern Recognition (CVPR)}. IEEE.

\bibitem[{Ginosar et~al.(2019{\natexlab{b}})Ginosar, Bar, Kohavi, Chan, Owens,
  and Malik}]{ginosar2019learning}
Ginosar, S.; Bar, A.; Kohavi, G.; Chan, C.; Owens, A.; and Malik, J.
  2019{\natexlab{b}}.
\newblock Learning individual styles of conversational gesture.
\newblock In \emph{Proceedings of the IEEE/CVF Conference on Computer Vision
  and Pattern Recognition}, 3497--3506.

\bibitem[{Goldin-Meadow(2005)}]{goldin2005hearing}
Goldin-Meadow, S. 2005.
\newblock \emph{Hearing gesture: How our hands help us think}.
\newblock Harvard University Press.

\bibitem[{Goldin-Meadow and McNeill(1999)}]{goldin1999role}
Goldin-Meadow, S.; and McNeill, D. 1999.
\newblock \emph{The role of gesture and mimetic representation in making
  language the province of speech}.
\newblock na.

\bibitem[{Hasegawa et~al.(2018)Hasegawa, Kaneko, Shirakawa, Sakuta, and
  Sumi}]{hasegawa2018evaluation}
Hasegawa, D.; Kaneko, N.; Shirakawa, S.; Sakuta, H.; and Sumi, K. 2018.
\newblock Evaluation of speech-to-gesture generation using bi-directional LSTM
  network.
\newblock In \emph{Proceedings of the 18th International Conference on
  Intelligent Virtual Agents}, 79--86.

\bibitem[{Heusel et~al.(2017)Heusel, Ramsauer, Unterthiner, Nessler, and
  Hochreiter}]{heusel2017gans}
Heusel, M.; Ramsauer, H.; Unterthiner, T.; Nessler, B.; and Hochreiter, S.
  2017.
\newblock Gans trained by a two time-scale update rule converge to a local nash
  equilibrium.
\newblock \emph{Advances in neural information processing systems}, 30.

\bibitem[{Ho, Jain, and Abbeel(2020)}]{ho2020denoising}
Ho, J.; Jain, A.; and Abbeel, P. 2020.
\newblock Denoising diffusion probabilistic models.
\newblock \emph{Advances in neural information processing systems}, 33:
  6840--6851.

\bibitem[{Huang and Mutlu(2012)}]{huang2012robot}
Huang, C.-M.; and Mutlu, B. 2012.
\newblock Robot behavior toolkit: generating effective social behaviors for
  robots.
\newblock In \emph{Proceedings of the seventh annual ACM/IEEE international
  conference on Human-Robot Interaction}, 25--32.

\bibitem[{Huang et~al.(2023)Huang, Huang, Yang, Ren, Liu, Li, Ye, Liu, Yin, and
  Zhao}]{huang2023make}
Huang, R.; Huang, J.; Yang, D.; Ren, Y.; Liu, L.; Li, M.; Ye, Z.; Liu, J.; Yin,
  X.; and Zhao, Z. 2023.
\newblock Make-an-audio: Text-to-audio generation with prompt-enhanced
  diffusion models.
\newblock \emph{arXiv preprint arXiv:2301.12661}.

\bibitem[{Ishi et~al.(2018)Ishi, Machiyashiki, Mikata, and
  Ishiguro}]{ishi2018speech}
Ishi, C.~T.; Machiyashiki, D.; Mikata, R.; and Ishiguro, H. 2018.
\newblock A speech-driven hand gesture generation method and evaluation in
  android robots.
\newblock \emph{IEEE Robotics and Automation Letters}, 3(4): 3757--3764.

\bibitem[{Kong et~al.(2020)Kong, Ping, Huang, Zhao, and
  Catanzaro}]{kong2020diffwave}
Kong, Z.; Ping, W.; Huang, J.; Zhao, K.; and Catanzaro, B. 2020.
\newblock Diffwave: A versatile diffusion model for audio synthesis.
\newblock \emph{arXiv preprint arXiv:2009.09761}.

\bibitem[{Kucherenko et~al.(2019)Kucherenko, Hasegawa, Henter, Kaneko, and
  Kjellstr{\"o}m}]{kucherenko2019analyzing}
Kucherenko, T.; Hasegawa, D.; Henter, G.~E.; Kaneko, N.; and Kjellstr{\"o}m, H.
  2019.
\newblock Analyzing input and output representations for speech-driven gesture
  generation.
\newblock In \emph{Proceedings of the 19th ACM International Conference on
  Intelligent Virtual Agents}, 97--104.

\bibitem[{Lee et~al.(2019)Lee, Yang, Liu, Wang, Lu, Yang, and
  Kautz}]{lee2019dancing}
Lee, H.-Y.; Yang, X.; Liu, M.-Y.; Wang, T.-C.; Lu, Y.-D.; Yang, M.-H.; and
  Kautz, J. 2019.
\newblock Dancing to music.
\newblock \emph{Advances in neural information processing systems}, 32.

\bibitem[{Li et~al.(2022)Li, Zhao, Zhelun, and Sheng}]{li2022danceformer}
Li, B.; Zhao, Y.; Zhelun, S.; and Sheng, L. 2022.
\newblock Danceformer: Music conditioned 3d dance generation with parametric
  motion transformer.
\newblock In \emph{Proceedings of the AAAI Conference on Artificial
  Intelligence}, volume~36, 1272--1279.

\bibitem[{Li et~al.(2021{\natexlab{a}})Li, Kang, Pei, Zhe, Zhang, He, and
  Bao}]{li2021audio2gestures}
Li, J.; Kang, D.; Pei, W.; Zhe, X.; Zhang, Y.; He, Z.; and Bao, L.
  2021{\natexlab{a}}.
\newblock Audio2gestures: Generating diverse gestures from speech audio with
  conditional variational autoencoders.
\newblock In \emph{Proceedings of the IEEE/CVF International Conference on
  Computer Vision}, 11293--11302.

\bibitem[{Li et~al.(2021{\natexlab{b}})Li, Yang, Ross, and Kanazawa}]{li2021ai}
Li, R.; Yang, S.; Ross, D.~A.; and Kanazawa, A. 2021{\natexlab{b}}.
\newblock Ai choreographer: Music conditioned 3d dance generation with aist++.
\newblock In \emph{Proceedings of the IEEE/CVF International Conference on
  Computer Vision}, 13401--13412.

\bibitem[{Liu et~al.(2022)Liu, Wu, Zhou, Xu, Qian, Lin, Zhou, Wu, Dai, and
  Zhou}]{liu2022learning}
Liu, X.; Wu, Q.; Zhou, H.; Xu, Y.; Qian, R.; Lin, X.; Zhou, X.; Wu, W.; Dai,
  B.; and Zhou, B. 2022.
\newblock Learning hierarchical cross-modal association for co-speech gesture
  generation.
\newblock In \emph{Proceedings of the IEEE/CVF Conference on Computer Vision
  and Pattern Recognition}, 10462--10472.

\bibitem[{Lu, Yoon, and Feng(2023)}]{lu2023co}
Lu, S.; Yoon, Y.; and Feng, A. 2023.
\newblock Co-Speech Gesture Synthesis using Discrete Gesture Token Learning.
\newblock \emph{arXiv preprint arXiv:2303.12822}.

\bibitem[{Lugmayr et~al.(2022)Lugmayr, Danelljan, Romero, Yu, Timofte, and
  Van~Gool}]{lugmayr2022repaint}
Lugmayr, A.; Danelljan, M.; Romero, A.; Yu, F.; Timofte, R.; and Van~Gool, L.
  2022.
\newblock Repaint: Inpainting using denoising diffusion probabilistic models.
\newblock In \emph{Proceedings of the IEEE/CVF Conference on Computer Vision
  and Pattern Recognition}, 11461--11471.

\bibitem[{Luo and Hu(2021)}]{luo2021diffusion}
Luo, S.; and Hu, W. 2021.
\newblock Diffusion probabilistic models for 3d point cloud generation.
\newblock In \emph{Proceedings of the IEEE/CVF Conference on Computer Vision
  and Pattern Recognition}, 2837--2845.

\bibitem[{Marsella et~al.(2013)Marsella, Xu, Lhommet, Feng, Scherer, and
  Shapiro}]{marsella2013virtual}
Marsella, S.; Xu, Y.; Lhommet, M.; Feng, A.; Scherer, S.; and Shapiro, A. 2013.
\newblock Virtual character performance from speech.
\newblock In \emph{Proceedings of the 12th ACM SIGGRAPH/Eurographics symposium
  on computer animation}, 25--35.

\bibitem[{Nyatsanga et~al.(2023)Nyatsanga, Kucherenko, Ahuja, Henter, and
  Neff}]{nyatsanga2023comprehensive}
Nyatsanga, S.; Kucherenko, T.; Ahuja, C.; Henter, G.~E.; and Neff, M. 2023.
\newblock A Comprehensive Review of Data-Driven Co-Speech Gesture Generation.
\newblock In \emph{Computer Graphics Forum}, volume~42, 569--596. Wiley Online
  Library.

\bibitem[{Pavlakos et~al.(2019)Pavlakos, Choutas, Ghorbani, Bolkart, Osman,
  Tzionas, and Black}]{SMPL-X:2019}
Pavlakos, G.; Choutas, V.; Ghorbani, N.; Bolkart, T.; Osman, A. A.~A.; Tzionas,
  D.; and Black, M.~J. 2019.
\newblock Expressive Body Capture: {3D} Hands, Face, and Body from a Single
  Image.
\newblock In \emph{Proceedings IEEE Conf. on Computer Vision and Pattern
  Recognition (CVPR)}, 10975--10985.

\bibitem[{Qi et~al.(2023)Qi, Liu, Li, Hou, Xin, and Yu}]{qi2023emotiongesture}
Qi, X.; Liu, C.; Li, L.; Hou, J.; Xin, H.; and Yu, X. 2023.
\newblock EmotionGesture: Audio-Driven Diverse Emotional Co-Speech 3D Gesture
  Generation.
\newblock \emph{arXiv preprint arXiv:2305.18891}.

\bibitem[{Razavi, Van~den Oord, and Vinyals(2019)}]{razavi2019generating}
Razavi, A.; Van~den Oord, A.; and Vinyals, O. 2019.
\newblock Generating diverse high-fidelity images with vq-vae-2.
\newblock \emph{Advances in neural information processing systems}, 32.

\bibitem[{Rombach et~al.(2022)Rombach, Blattmann, Lorenz, Esser, and
  Ommer}]{rombach2022high}
Rombach, R.; Blattmann, A.; Lorenz, D.; Esser, P.; and Ommer, B. 2022.
\newblock High-resolution image synthesis with latent diffusion models.
\newblock In \emph{Proceedings of the IEEE/CVF conference on computer vision
  and pattern recognition}, 10684--10695.

\bibitem[{Van Den~Oord, Vinyals et~al.(2017)}]{van2017neural}
Van Den~Oord, A.; Vinyals, O.; et~al. 2017.
\newblock Neural discrete representation learning.
\newblock \emph{Advances in neural information processing systems}, 30.

\bibitem[{Yang et~al.(2023)Yang, Wu, Li, Zhang, Hao, Bao, Cheng, and
  Xiao}]{yang2023diffusestylegesture}
Yang, S.; Wu, Z.; Li, M.; Zhang, Z.; Hao, L.; Bao, W.; Cheng, M.; and Xiao, L.
  2023.
\newblock DiffuseStyleGesture: Stylized Audio-Driven Co-Speech Gesture
  Generation with Diffusion Models.
\newblock \emph{arXiv preprint arXiv:2305.04919}.

\bibitem[{Yang, Yang, and Hodgins(2020)}]{yang2020statistics}
Yang, Y.; Yang, J.; and Hodgins, J. 2020.
\newblock Statistics-based motion synthesis for social conversations.
\newblock In \emph{Computer Graphics Forum}, volume~39, 201--212. Wiley Online
  Library.

\bibitem[{Ye et~al.(2023)Ye, Jia, Wu, Huang, Sun, and Xing}]{ye2023salient}
Ye, Z.; Jia, J.; Wu, H.; Huang, S.; Sun, S.; and Xing, J. 2023.
\newblock Salient Co-Speech Gesture Synthesizing with Discrete Motion
  Representation.
\newblock In \emph{ICASSP 2023-2023 IEEE International Conference on Acoustics,
  Speech and Signal Processing (ICASSP)}, 1--5. IEEE.

\bibitem[{Yi et~al.(2023)Yi, Liang, Liu, Cao, Wen, Bolkart, Tao, and
  Black}]{yi2023generating}
Yi, H.; Liang, H.; Liu, Y.; Cao, Q.; Wen, Y.; Bolkart, T.; Tao, D.; and Black,
  M.~J. 2023.
\newblock Generating holistic 3d human motion from speech.
\newblock In \emph{Proceedings of the IEEE/CVF Conference on Computer Vision
  and Pattern Recognition}, 469--480.

\bibitem[{Yoon et~al.(2020)Yoon, Cha, Lee, Jang, Lee, Kim, and
  Lee}]{yoon2020speech}
Yoon, Y.; Cha, B.; Lee, J.-H.; Jang, M.; Lee, J.; Kim, J.; and Lee, G. 2020.
\newblock Speech gesture generation from the trimodal context of text, audio,
  and speaker identity.
\newblock \emph{ACM Transactions on Graphics (TOG)}, 39(6): 1--16.

\bibitem[{Yoon et~al.(2019)Yoon, Ko, Jang, Lee, Kim, and Lee}]{yoon2019robots}
Yoon, Y.; Ko, W.-R.; Jang, M.; Lee, J.; Kim, J.; and Lee, G. 2019.
\newblock Robots learn social skills: End-to-end learning of co-speech gesture
  generation for humanoid robots.
\newblock In \emph{2019 International Conference on Robotics and Automation
  (ICRA)}, 4303--4309. IEEE.

\bibitem[{Zhang et~al.(2023)Zhang, Zhang, Cun, Huang, Zhang, Zhao, Lu, and
  Shen}]{zhang2023generating}
Zhang, J.; Zhang, Y.; Cun, X.; Huang, S.; Zhang, Y.; Zhao, H.; Lu, H.; and
  Shen, X. 2023.
\newblock T2M-GPT: Generating Human Motion from Textual Descriptions with
  Discrete Representations.
\newblock In \emph{Proceedings of the IEEE/CVF Conference on Computer Vision
  and Pattern Recognition (CVPR)}.

\bibitem[{Zhu et~al.(2023)Zhu, Liu, Liu, Qian, Liu, and Yu}]{zhu2023taming}
Zhu, L.; Liu, X.; Liu, X.; Qian, R.; Liu, Z.; and Yu, L. 2023.
\newblock Taming Diffusion Models for Audio-Driven Co-Speech Gesture
  Generation.
\newblock In \emph{Proceedings of the IEEE/CVF Conference on Computer Vision
  and Pattern Recognition}, 10544--10553.

\end{thebibliography}
\newpage

\section{Appendix}

In the supplementary material, we 1) provide more model details of C2G2, 2) give the detailed description of evaluation metrics 3) show more experimental details on C2G2.

\section{Model Details}
\subsection{C2G2 Model Architecture}
We provide the detailed model architecture of C2G2 as shown in Table \ref{arch}.
We only give the description of the encoder in temporal-aware VQVAE. 
The decoder has symmetric structure with encoder but with different number of Trans-Blocks (4 in encoder, 2 in decoder).

\subsection{In-betweening Editing Details}
In-betweening generation allows to generate gestures between pre-condition frames and post-condition frames. 
It can be further used to add/replace short movements to any position of a long sequence.
Specifically, in the user study, i.e., the middle clip (34 frames in total) generation, repainting strategy applies in the generation of the last clip where last 10 frames are post-condition frames.
However, only 6 frames before the post-condition frames are used as the transition part and are generated using repainting strategy, while the former frames are still sampled from noise.
\begin{table}[t]
\centering
\renewcommand{\arraystretch}{1.25}
\resizebox{\linewidth}{!}{
\begin{tabular}{c|c|c} 
\hline 
\textbf{Block}&\textbf{Shape}&\textbf{Operation}\\ \hline \hline
\multicolumn{3}{c}{\textbf{VQ-VAE Encoder}} \\
\hline
input&34*126&\\
\hline 
LIN(126,128)&34*128&Linear,LN,LeakyReLU\\
\hline 
ResLIN(128,128)&34*128&Linear,LN,Linear,LN\\
\hline 
ResLIN(128,256)&34*256&Linear,LN,Linear,LN\\
\hline
ResLIN(256,512)&34*512&Linear,LN,Linear,LN\\
\hline
ResLIN(512,256)&34*256&Linear,LN,Linear,LN\\
\hline
ResLIN(256,128)&34*128&Linear,LN,Linear,LN\\
\hline
LIN(128,128)&34*128&Linear,LN,LeakyReLU\\
\hline
Position-Encoder&34*128&Positional Embedding\\
\hline
Trans-Block *4&34*128&Self-attention,FFN\\
\hline
LIN(128,128)&34*128&Linear,LN,LeakyReLU\\
\hline
Quant-Conv(128,128)&34*128&Conv1d\\ \hline \hline 
\multicolumn{3}{c}{\textbf{Audio Encoder}} \\ \hline
Convolution layer1 &16*7891&Conv1d,BN,LeakyReLU\\
\hline
Convolution layer2&32*1313&Conv1d,BN,LeakyReLU\\
\hline
Convolution layer3&64*217&Conv1d,BN,LeakyReLU\\
\hline
Convolution layer4&32*34&Conv1d\\ \hline \hline
\multicolumn{3}{c}{\textbf{Latent Diffusion Model}} \\ \hline

LIN-In&34*256&Linear,LN,LeakyReLU\\
\hline
Trans-Block*8&34*256&self-attention, FFN\\
\hline
LIN-Out-1&34*128&Linear,LN,LeakyReLU\\
\hline
LINr-Out-2&34*128&Linear\\ \hline
\end{tabular}}
\caption{Detailed architecture of C2G2.}
\label{arch}
\end{table}



\section{Evaluation Metrics}
In the evaluation, we adopt three different metrics to measure the naturalness and correctness of the generated gesture sequences. 

\noindent \textbf{Frechet Gesture Distance (FGD). } Similar as Frechet Inception Distance (FID) \cite{heusel2017gans}, we use an autoencoder pretrained on both datasets to compute the distance between the generated gesture sequences and the real ones in the latent feature space. 
FGD is the metric best describing the synthesis quality of the generated sequences. 

\noindent \textbf{Beat Consistency Score (BC). } 
To measure the correlation of generated gesture sequences and the corresponding audio sequences, we use Beat Consistency (BC) \cite{li2022danceformer,li2021ai}.
To obtain BC, we need to first calculate the mean absolute angle change \cite{liu2022learning}, followed by the angle change rate.
Motion beats are then estimated by using such a change of included angle between bones.
Following \cite{ellis2007beat} , we detect the audio beats by onset strength \cite{li2022danceformer}.
Finally, we calculate the average distance between the audio beat and its nearest motion beat to serve as Beat Consistency Score.

\noindent \textbf{Diversity. } The metric measures the variation of generated gestures compared with corresponding inputs \cite{lee2019dancing}. 
Specifically, it is calculated as the mean distance of a set of generated sequences and its randomly shuffled variant in the latent feature space that is described in FGD.
In practice, we randomly select 500 different gesture sequences and then calculate the average feature distance accordingly.
\begin{figure*}[t]
\centering
\resizebox{\linewidth}{!}{
\includegraphics[width=1.5\columnwidth]{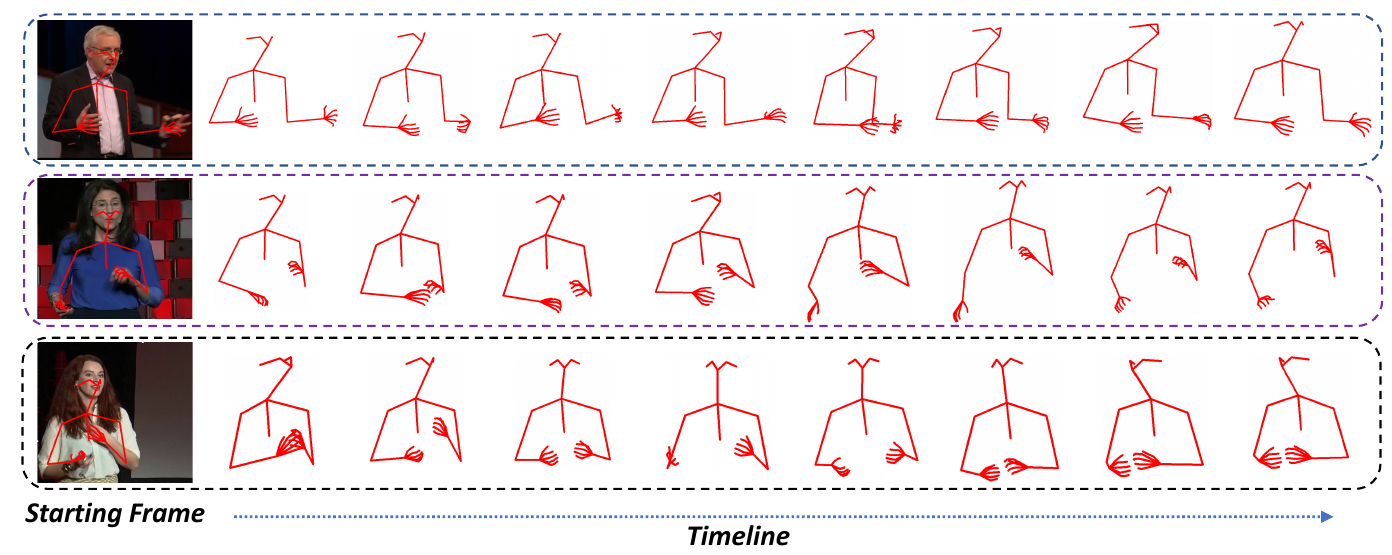} }
\caption{Visualization of speaker conditional generation.}
\label{vqvae}
\vspace{-0.5cm}
\label{spk_seq}
\end{figure*}

\section{More Experimental Results}
\subsection{Ablation Study on Codebook Shape}
To analyze the effect of the codebook shape on the VQ-VAE performance, we conduct experiments on various settings with different codebook size and code channels (dimensionality). 
We test 6 settings in total with different codebook sizes and code channels, where the results are show in Table \ref{codebook}. 
The result reveal that, in the setting of $(1024, 128)$, the VQ-VAE model achieves the best performance as reported in the main paper. 
We do not consider more channels as this leads to the difficulty and high cost of diffusion training, which hinders the practical usage.

\begin{table}
\centering
\renewcommand{\arraystretch}{1.1}
\resizebox{\linewidth}{!}{
\begin{tabular}{ccccc} 
\hline
\multicolumn{1}{c}{\multirow{2}{*}{\textbf{Methods}}} & \multicolumn{2}{c}{\textbf{Shape}} & \multicolumn{2}{c}{\textbf{TED Expressive}} \\ \cline{2-3} \cline{4-5}
\multicolumn{1}{c}{}                                   &Size&Channel& FGD $\downarrow$& Diversity $\uparrow$\\        \hline
Ground-Truth&&&0&178.827\\
\hline
VQ-VAE &512&64&0.854 & 178.245\\
VQ-VAE &512&128&0.849&178.888\\
VQ-VAE &512&256&0.588&179.366\\
VQ-VAE&1024&64&0.560&\textbf{180.261}\\
VQ-VAE &1024&128&\textbf{0.402}&{179.743}\\
VQ-VAE&1024&256&0.526&178.918\\
\hline 
\end{tabular}}
\caption{Ablation study results of different codebook shapes in terms of codebook size and code channels.}
\label{codebook}
\end{table}

\subsection{Real-length Input Training Pipeline}
In the paper, the VQ-VAE is trained using gesture sequences represented by unit vectors and then the speaker information is then introduced through a specifically designed speaker related decoder.
The reason of not directly using real-length vectors is because this will lead to unstable training (not converging) and thus jeopardize the generated results.
To verify this, we conduct experiments of directly reconstructing real-length gestures by VQ-VAE without using the SRD.
As shown in table \ref{conv}, VQ-VAE trained with the real-lenght vectors faces serious mode collapse issue and poor reconstruction results on various settings of codebook size and code channel. 



\begin{table}
\centering
\renewcommand{\arraystretch}{1.1}
\resizebox{\linewidth}{!}{
\begin{tabular}{ccccc} 
\hline 
\multicolumn{1}{c}{\multirow{2}{*}{\textbf{Methods}}} & \multicolumn{2}{c}{\textbf{Shape}} & \multicolumn{2}{c}{\textbf{TED Expressive}} \\ \cline{2-5}
\multicolumn{1}{c}{}   &Size&Channel& FGD $\downarrow$ & Rec-loss $\downarrow$ \\
\hline
Ground Truth & & &0.000&0.000\\
\hline
VQ-VAE$_{rl}$ &1024&64&281.65&0.021\\
VQ-VAE$_{rl}$ &1024&128&283.40&0.019\\
VQ-VAE$_{rl}$ &1024&256&280.01&0.021\\
VQ-VAE$_{rl}$&2048&128&282.22&0.019\\
VQ-VAE$_{rl}$&2048&256&283.66&0.019\\
VQ-VAE$_{ours}$ &1024&128&\textbf{0.341}&\textbf{0.005}\\
\hline 
\end{tabular}}
\caption{Results of VQ-VAE trained by real-length vectors on various codebook settings.}
\label{conv}
\end{table}

\begin{figure*}[t]
\centering
\includegraphics[width=1.8\columnwidth]{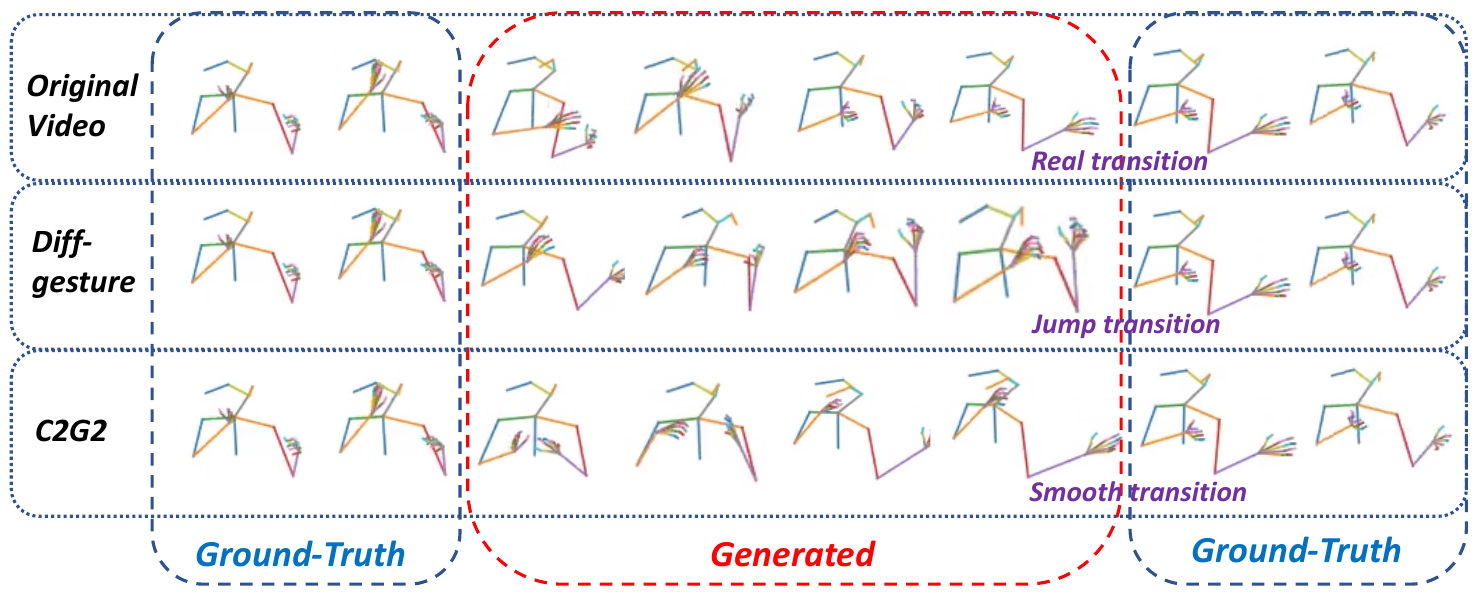} 
\caption{Visualization of in-betweening generation.}
\label{vqvae}
\label{inbetweening}
\end{figure*}



\subsection{Subjective Evaluation Protocol}
For the user study, we design a detailed scoring protocol for the subjective evaluation.
Rating score is from 1 to 5 with 1 as the worst and 5 as the best. 
We give the detailed descriptions for score 1, 3 and 5.

\noindent \emph{\textbf{Naturalness:}}
\begin{itemize}
\item 1 point - generated gesture sequence is totally unnatural, and may contain serious frozen, distortion and shaking movements. 
\item 3 point - generated gesture sequence is partially natural, part of the poses may include generated traces of unreasonable movements or sight frozen.
\item 5 point - generated gesture sequence follows the usual movement habits and margins of real humans without noticeable issues.
\end{itemize}

\noindent \emph{\textbf{Smoothness:}}
\begin{itemize}
\item 1 point - result has observe inconsistency across temporal sequence and body parts. Rapid changes usually happen between frames.
\item 3 point - body skeleton is mainly temporal consistent with satisfied smoothness. Inconsistency may occur in hands. 
\item 5 point - overall result has strong temporal consistency and is smooth both in body and hands.
\end{itemize}

\noindent \emph{\textbf{Synchrony:}}
\begin{itemize}
\item 1 point - generated gesture sequence has no related alignment for the speech content and beat.
\item 3 point - generated gesture sequence is synchronized with the input speech in several parts, and correlation could be observed.
\item 5 point - generated gesture sequence is highly aligned with the accompanied speech, and no unrelated gesture is observed in the whole sequence.
\end{itemize}

\subsection{Visualization Results}
We provide more visualization results in this section. In figure \ref{spk_seq}, we show the generated results of 3 different speakers. In figure \ref{inbetweening}, we show the smooth generation results of the in-betweening case. More video samples can be found in our demo page \url{https://c2g2-gesture.github.io/c2_gesture/}.

\end{document}